\documentclass[11pt]{article}

\usepackage[final]{acl}

\usepackage{microtype}
\usepackage{ifluatex}
\usepackage{amsmath}
\usepackage{multirow}
\usepackage{enumitem}
\usepackage{float}
\usepackage{booktabs}
\usepackage{adjustbox}
\usepackage[table]{xcolor}
\usepackage[english,bidi=default]{babel} 
\babelfont{rm}{TeXGyreTermesX} 

\ifluatex
\newfontfamily\hausafont{DejaVu Serif}
\else
\newcommand{\hausafont}[1]{#1} 
\fi

\usepackage{newunicodechar}
\newunicodechar{ɓ}{{\hausafont\fontsize{9pt}{10pt}\selectfont ɓ}}
\newunicodechar{ɗ}{{\hausafont\fontsize{9pt}{10pt}\selectfont ɗ}}
\newunicodechar{ƙ}{{\hausafont\fontsize{9pt}{10pt}\selectfont ƙ}}
\newunicodechar{ƴ}{{\hausafont\fontsize{9pt}{10pt}\selectfont ƴ}}

\title{Automatic Correction of Writing Anomalies in Hausa Texts}

\author{Ahmad Mustapha Wali \and Sergiu Nisioi\thanks{Corresponding author.} \\
         Human Language Technologies Research Center \\ Faculty of Mathematics and Computer Science \\ University of Bucharest \\ \texttt{ahmadmwali@gmail.com, sergiu.nisioi@unibuc.ro}}

\begin{document}

\maketitle
\begin{abstract}
Hausa texts are often characterized by writing anomalies, such as incorrect character substitutions and spacing errors, which sometimes hinder natural language processing (NLP) applications. This paper presents an approach to automatically correct anomalies by finetuning transformer-based models. Using a corpus gathered from several public sources, we create a large-scale parallel dataset of over 400,000 noisy-clean Hausa sentence pairs by introducing synthetically generated noise to mimic realistic writing errors. In addition, we finetune several multilingual and African language models, including M2M100, AfriTeVA, \texttt{NCAIR1/N-ATLaS}, \texttt{UBC-NLP/cheetah-base}, and other variants of BART and T5 for this correction task. Our experimental results demonstrate that models such as M2M100 achieve state-of-the-art results despite their smaller size and distinct pretraining, and that correcting errors can have a significant impact in improving downstream tasks such as text classification, machine translation, question answering, and LLM prompting in general. This research provides a methodology, a publicly available dataset, and a comparison of models to improve Hausa text quality, thereby advancing NLP capabilities for the language and offering transferable insights for other low-resource languages.
\end{abstract}

\section{Introduction} 
Hausa (\texttt{ha}, ISO 639-1) belongs to the Chadic branch of the Afro-Asiatic language family and is widely spoken in West Africa. The language has deep historical roots and is connected to other major Afro-Asiatic languages (from the Semitic, Cushitic, and Berber families) that span regions of Africa and West Asia \cite{1}. As a major regional language, Hausa has a large speaker base of approximately 80 million people, primarily residing in West Africa \cite{2}. The majority of Hausa speakers in Nigeria and Niger are first-language users \cite{2}. Beyond these areas, Hausa functions as a lingua franca in regions where it is not the native language \cite{2, 3}, being a key language for inter-group communication and regional integration across Nigeria's Middle Belt, northern Ghana, and the Benin Republic.

The increased use of Hausa in digital communication, particularly on social media, has raised new concerns about orthographic variations and the use of informal languages \cite{5}. The prevalence of writing errors in digitally available texts further exacerbates the lack of high-quality data for Hausa NLP. These irregularities can be characterized as either character substitutions, where certain Hausa characters (ɓ, ɗ, ƙ, and ƴ) are replaced with standard Latin alphabet, or spacing issues, which involve both the removal and addition of spaces between words. Although these writing anomalies are easily disambiguated by humans, they pose significant challenges to NLP models. 

For example, the sentence ``abincin ba shi da daɗi'' (Eng. \textit{the food is not delicious}) is commonly written as “abincin bashi da daɗi,” (Eng. \textit{food bought on credit is delicious}) a shift in semantics that completely affects the meaning of the sentence. Character swaps, such as “Wannan ƴa ta ce” (Eng. \textit{this is my daughter}) becoming “Wannan ya ta ce” (Eng. \textit{this is my elder sister}), introduce confusion and noise into the data. 
Although model architecture and training methods are important, the performance of NLP systems also depends on the quality of the training data \cite{6}. This factor is especially relevant for low-resource languages such as Hausa, where data availability is limited.

Our work fills several gaps for Hausa NLP, firstly by building a large synthetic dataset of writing anomalies using a small seed of social-media data, secondly by building the first set of tools for correcting writing anomalies in Hausa texts based on transformer architectures \cite{6, 24}, and thirdly by evaluating the impact of correcting such errors in downstream tasks. To summarize our contributions, we provide:
\begin{itemize}
    \item An analysis and categorization of common writing anomalies prevalent in digital Hausa texts.
    \item A new dataset consisting of over \~400,000 samples, constructed by synthetically generating noise, mimicking common Hausa writing errors, thereby creating noisy-clean parallel pairs\footnote{Released under CC-BY-4.0 license.} suitable for text correction tasks and broader Hausa NLP research.
    \item We train and evaluate several Transformer-based models for the automatic detection and correction of these identified Hausa writing anomalies.
    \item We present a quantitative evaluation of text quality improvements achieved by our correction models, measured using standard text evaluation metrics and assessed through their impact on downstream tasks.
\end{itemize}

\section{Related Work}

\subsection*{Challenges in Low-Resource Language Processing}
Data scarcity is a major challenge in low-resource NLP, and low-resource languages have a distinct disadvantage because they typically lack the digitized texts, annotated corpora, and parallel data required for machine translation \cite{lusito2023text}. This lack of training data directly decreases the ability of an NLP model to learn linguistic patterns and generalize efficiently, lowering overall performance \cite{6, 10}. In addition, reduced digital literacy rates, as well as the strong oral traditions associated with certain low-resource languages, further limit corpus generation and the representativeness of such data \cite{11, adelani-etal-2025-generative}.

Numerous low-resource languages have complex morphology and lexicon systems, which pose additional challenges \cite{13, oncevay-etal-2022-schaman}. In particular, Hausa is an agglutinative language, creating challenges less common in high-resource contexts such as English or Romance languages \cite{15, Uwaezuoke_Anachunam_2023}. 

Furthermore, low-resource languages have limited accessible tools and resources, unlike high-resource languages, which have an advantage due to the extensive ecosystems of pretrained language models, comprehensive NLP libraries, and easily accessible infrastructure, including standard keyboards and encoding techniques. Materials in low-resource languages often require researchers to allocate time and funds judiciously to establish tools and resources from scratch \cite{11}.

Orthographic variations and informal language use, especially in digital texts from low-resource environments \cite{16}, further exacerbate these issues. Consequently, these diverse challenges directly impact the effectiveness of NLP tasks for low-resource languages. For machine translation, for example, the lack of parallel corpora leads to lower translation quality compared to pairs of high-resource languages \cite{18}. Similarly, a lack of annotated data and natural language complexity restrict operations such as part-of-speech tagging, parsing, and named entity recognition, decreasing precision and robustness \cite{18, 19}.

\begin{figure*}[!htb]
    \centering
    \includegraphics[width=1.5\columnwidth]{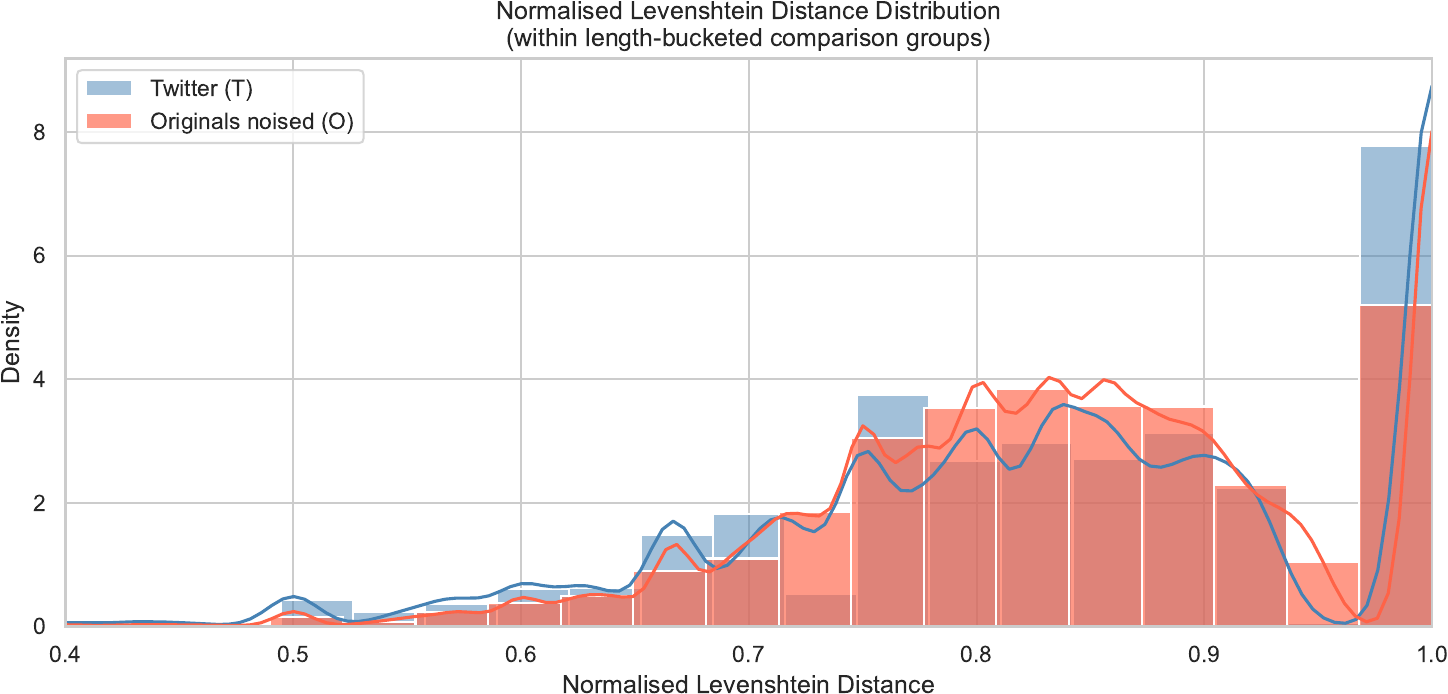} 
    \caption{Our dataset's characteristics are calibrated so that its normalized Levenshtein distance distribution resembles that of the naturally-noisy Twitter dataset. This process results in a parallel corpus of noisy-to-clean sentence pairs.
    Normalized Levenshtein Distance Distribution between the twitter dataset and the synthetically-generated noisy dataset calibrate with the parameters from Table 2.}
    \label{fig:levenshtein_dist}
\end{figure*}

\subsection*{Hausa Natural Language Processing: Status and Specific Challenges}
NLP tools for Hausa have lately shown considerable growth \cite{bashir2017automatic,21,akinfaderin-2020-hausamt,rakhmanov-schlippe-2022-sentiment,parida-etal-2023-havqa,ahmad-etal-2024-generative,alabi-etal-2025-afridoc,sani-etal-2025-investigating,uemura-etal-2026-afrimteb} owing in part to the recent series of AfricaNLP Workshops \cite{africanlp-ws-2025-1,africanlp-2026-main} and the sustained effort of the Masakhane community \cite{hausaner,dione2023masakhapos,muhammad-etal-2025-hausanlp}. Despite the recent growth of resources, challenges still exist. Hausa's morphological system provides specific computational problems by combining concatenative and non-concatenative processes \cite{20}. This intricacy is notable in its verbal system, where derivation from a single root can create several forms that need robust morphological analytic methods. For example, the root word ``karya'' (\textit{to break}) has several forms, including karyawa (\textit{breaking}), karye (\textit{broken}), kakkarye (\textit{broken several times}), kari (\textit{breaking in noun form}), karyayye (\textit{a broken object (m)}) and so on. Hausa's interaction of tone and meaning adds yet another level of complexity that existing NLP systems are sometimes unable to comprehend \cite{20, 5}, such as how the word ``baba'' can mean any of the words \textit{father}, \textit{mother}, or \textit{pal} depending on the intonation used.

Transformer-based models have opened new pathways for Hausa NLP, with experiments using multilingual African models \cite{belay-etal-2025-afroxlmr} demonstrating potential applications in named entity recognition (NER) and sentiment analysis, among others \cite{muhammad-etal-2022-naijasenti, 24, rakhmanov-schlippe-2022-sentiment}. However, owing to inadequate representation in pretraining data, their performance in Hausa consistently lags behind that of high-resource languages \cite{5}. Limitations in computational resources and the lack of high-quality training data \cite{25} exacerbate the challenges of fine-tuning these models.

These technical challenges are amplified by broader infrastructural and institutional constraints. While many Nigerian and Nigerien institutions provide computer science and linguistics courses, specialized education in computational Hausa language processing is still rare \cite{5}. Some of these issues are starting to be addressed by recent community-led, multinational cooperative projects. HausaNLP \cite{emezue2023africanstopwordsprojectcurating,ahmad2025exploringculturalnuancesemotion,hussen2025statelargelanguagemodels} provides a platform for coordinating research, developing models and datasets, and sharing resources among institutions focused on Hausa and other African languages.

\section{Data Collection and Preparation}
The foundation of our work is a large Hausa corpus aggregated from multiple publicly available datasets. 
The specific datasets incorporated include Hausa Wikipedia (a crawl of the existing data), Wikimedia Hausa, and a collection of other miscellaneous Hausa texts.
Furthermore, several texts of different genres from the MultiplEYE project \cite{kasper2026multipleye,nisioi2026multipleye} have been translated into Hausa following the official project guidelines \cite{meyeguide}. These texts are included as additional out-of-domain (\textbf{OOD}) data, since they have not been previously published online and can be used for a fair comparison of models. The texts are multi-genre and cover two argumentative texts from the Programme for International Student Assessment (PISA), two institutional texts (the Declaration of Human Rights and a text from the European Commission), seven literary texts, and two pop-science texts. In total, we include 10,000 words of high-quality translated texts into Hausa. 

After merging these sources, the texts are cleaned by removing common digital artifacts such as hashtags, non-breaking space markers (``NBSP''), and citation patterns. The merged and cleaned text was then segmented into sentences. This process yielded over 400,000 clean Hausa sentences, distributed as follows: Hausa Wikipedia (160k), Wikimedia Hausa (214k), Other Texts (29k). The validation and training sets each contain 5,000 sentences.

Because there is no readily available data annotated with errors for Hausa, we are facing a chicken-and-egg problem. As such, the majority of Hausa errors are not to be found in (quasi-) official sources such as our dataset, but in more informal texts published on social media.

We use the NaijaSenti dataset of approximately 10,000 Hausa tweets \cite{muhammad-etal-2022-naijasenti,Muhammad2023AfriSentiAT}, which typically contain the types of problems we are seeking to address. Potential misspellings are first identified using a lexicon-based approach: tokens absent from a standardized Hausa word list are flagged as out-of-vocabulary (OOV). For each OOV token, we compute its similarity to in-vocabulary items using a normalized Levenshtein distance. 
Tokens longer than two characters are grouped into length-based buckets $[\text{len}(x) - 1,\, \text{len}(x) + 1]$, and within each bucket, normalized Levenshtein distances are computed for all word pairs. Using DBSCAN (\texttt{eps}=0.4, \texttt{min\_samples}=2) with the precomputed distance matrix, clusters of similar misspellings are identified while ignoring singleton (noise) clusters. From these results, we collect distributions of normalized edit distances and cluster sizes, which are visualized as histograms for both datasets (see \autoref{fig:levenshtein_dist}). We derive a distribution of error types, distinguishing between character substitutions, insertions, deletions, and spacing errors. The types of noise introduced include: 1. incorrect character substitution by replacing Hausa-specific hooked letters (ɓ, ɗ, ƙ, ƴ, and their uppercase counterparts) with their plain English alphabet equivalents (b, d, k, y); 2. random spacing errors and space removal errors, e.g., merging adjacent words by removing the intervening space; 3. random character deletion, random character duplication, random substitution of some characters with other characters; 4. chunk deletion and insertion, operating on word-level 2-character. These changes are irreversible using a rule-based system.
The probabilities of these operations are selected randomly, before being subsequently visualized against the normalized distances of the Twitter dataset. The process is repeated until the normalized distances of the synthetic dataset match the distances extracted using the Twitter dataset. To assess the reliability of the synthetically generated noise, a native speaker manually inspects the generated content across different text genres in a non-exhaustive manner.
\begin{table}[!htb]
    \centering
    \begin{tabular}{ll}
        \toprule
        Noise Type & Probability \\
        \midrule
        random\_spacing & 0.02 \\
        remove\_spaces & 0.15 \\
        incorrect\_characters & 0.02 \\
        delete\_characters & 0.005 \\
        duplicate\_characters & 0.01 \\
        substitute\_characters & 0.001 \\
        transpose\_characters & 0.01 \\
        delete\_chunk & 0.0015 \\
        insert\_chunk & 0.001 \\
        \bottomrule
    \end{tabular}
    \caption{Character-Level Noise Configuration.}
    \label{tab:noise_proba}
\end{table}
The derived probabilities of these errors are rendered in \autoref{tab:noise_proba} and are used to generate \textbf{synthetic alterations} to the original dataset to reflect human-like anomalies. Additionally, we compare the distribution of writing errors in the synthetic dataset using  the Jensen--Shannon (JS) distance between two discrete distributions $T$ (twitter) and $O$ (originals), defined as
\[
JS(T \,\|\, O) = \sqrt{\tfrac{1}{2} D_{KL}(T \,\|\, M) + \tfrac{1}{2} D_{KL}(O \,\|\, M)}
\]
where $ M = \tfrac{1}{2}(T + O)$ and $D_{KL}$ denotes the Kullback--Leibler divergence. 
The computed JS distance of $0.14$ indicating a good match between the empirical and synthetic distance distributions. 


\begin{table*}[!ht]
\centering
\begin{tabular}{@{}lccccc@{}}
\toprule
Model & BLEU $\uparrow$ & METEOR $\uparrow$ & F1 $\uparrow$ & WER $\downarrow$ & CER $\downarrow$ \\
\midrule
Copy Baseline & 0.293 & 0.654 & 0.554 & 0.501 & 0.082 \\
LSTM Baseline & 0.156 & 0.389 & 0.757 & 0.635 & 0.534 \\
GPT-5.1 Baseline & 0.595 & 0.743 & 0.773 & 0.271 & 0.067  \\
\midrule
\rowcolor{gray!15}NCAIR1/N-ATLaS (8B)  & 0.886 & 0.929 & 0.936 & 0.079 & 0.013 \\
\rowcolor{gray!15}M2M100 (418M) & 0.853 & 0.932 & 0.933 & 0.079 & 0.017 \\
UBC-NLP/cheetah-base (1.2B) & 0.834 & 0.924 & 0.926 & 0.086 & 0.028 \\
AfriTeVA-Base (580M) & 0.754 & 0.880 & 0.884 & 0.135 & 0.045 \\
AfriTeVA-Small (229M) & 0.656 & 0.823 & 0.835 & 0.203 & 0.075 \\
AfriMBART (680M) & 0.789 & 0.853 & 0.855 & 1.862 & 1.825 \\
mBART-Large-50 (610M) & 0.413 & 0.661 & 0.730 & 0.352 & 0.090 \\
mT5-Base (580M) & 0.656 & 0.823 & 0.841 & 0.267 & 0.158\\
\bottomrule
\end{tabular}%
\caption{Performance of various transformer models on the Hausa text correction test set. Best-performing models are highlighted in gray. Lower is better for WER and CER; higher is better for others. The M2M100 adapted for the text normalization task achieves strong results, better than models multilingually adapted to African languages. The only model that is marginally better is a finetuned 8B model.
GPT-5.1 under-performs for Hausa text normalization and is comparable to the baseline of doing no transformation to the data (first row).}
\label{tab:results}
\end{table*}

\section{Automatic Correction Models}
The automatic correction task is framed as translating noisy Hausa sentence into their corrected standard forms. We investigate a variety of multilingual and African transformer models by fine-tuning them on the synthetically generated noisy-clean parallel Hausa dataset.
We adopt the following pretrained models that have been exposed to Hausa.
\begin{itemize}
    \item \textbf{M2M100} \cite{fan2020englishcentricmultilingualmachinetranslation} is a seq2seq massively multilingual model, we use the 418M parameter version. These models are designed for many-to-many multilingual translation and can be adapted for monolingual correction tasks.
    \item \textbf{AfriTeVA}: we utilize both the ``base'', ``small'' and ``large'' versions of AfriTeVA \cite{jude-ogundepo-etal-2022-AfriTeVA}, a seq2seq model pretrained on a diverse set of African languages, including Hausa. 
    \item \textbf{mT5} \cite{xue2021mt5} is a multilingual T5 transformer including Hausa in its 101 pretraining languages.
    \item \textbf{AfriMBART} \cite{adelani-etal-2022-thousand} a multilingual BART model \cite{tang2020multilingual} trained for machine translation of 16 African languages. 
    \item \textbf{UBC-NLP/cheetah-base} \cite{adebara-etal-2024-cheetah} a massively multilingual T5 language model for 517 African languages and language varieties. 
    \item \textbf{NCAIR1/N-ATLaS} \cite{awagptv1_2025} is a fine-tuned multilingual language model based on Llama-3 8B designed to support African languages, including Hausa, Igbo, and Yoruba alongside English.
\end{itemize}

These models are fine-tuned using the noisy Hausa sentences as input and the corresponding clean sentences as the target output. The N-ATLaS model is fully finetuned for two epochs using the Axolotl \cite{axolotl} framework and a base instruction copied verbatim in \autoref{sec:llm_sft}. Additionally, we employ three more baselines: 1. a copy baseline that simply copies the noisy texts with no operation; 2. an LSTM-based sequence-to-sequence model; and 3. GPT-5.1 prompted three times (see \autoref{sec:LLM_correction_prompt}) with default parameters, average scores are reported $\pm 0.001$. Details on hyperparameters for all the other models are available in \autoref{sec:app_training} and in the released code.
\begin{table*}[h]
\centering
\begin{tabular}{lccccc}
\toprule
Model & BLEU $\uparrow$ & METEOR $\uparrow$ & F1 $\uparrow$ & WER $\downarrow$ & CER $\downarrow$ \\
\midrule
\rowcolor{gray!15}NCAIR1/N-ATLaS (8B)  & 0.808 & 0.879 & 0.892 & 0.169 & 0.089 \\
\rowcolor{gray!15}M2M100 (418M)        & 0.792 & 0.885 & 0.896 & 0.117 & 0.024 \\
UBC-NLP/cheetah-base & 0.740 & 0.860 & 0.871 & 0.174 & 0.080 \\
AfriTeVA-Base (580M) & 0.640 & 0.817 & 0.847 & 0.625 & 0.512 \\
AfriTeVA-Small (229M)& 0.547 & 0.770 & 0.820 & 1.039 & 0.874 \\
AfriMBART (680M)     & 0.632 & 0.714 & 0.720 & 5.991 & 6.423 \\
\bottomrule
\end{tabular}
\caption{Evaluation results on Hausa noisy-to-clean correction for out-of-domain data. The values are overall smaller than for the in-domain data \autoref{tab:results}, however, the hierarchy of the models is preserved - M2M and \texttt{NCAIR1/N-ATLaS} being the best-performing models.}
\label{tab:ood_res}
\end{table*}

\subsection{Automatic Evaluation Metrics}
Character Error Rate \textbf{(CER)} \cite{36, 37} measures the minimum number of edit operations (insertions, deletions, substitutions) required to transform the predicted output into the reference text, divided by the length of the reference text. 

Word Error Rate \textbf{(WER)} calculates the number of word-level edits needed to transform the corrected sentence into the reference sentence, divided by the number of words in the reference sentence. The metric is useful when errors predominantly affect word boundaries or word choice \cite{38}.  

\textbf{BLEU}\footnote{Signature: \path{nrefs:1|case:mixed|eff:no|tok:13a|smooth:exp|version:2.4.3}} from SacreBleu \cite{post-2018-call} measures n-gram precision overlap between the generated text and the reference.

\textbf{METEOR} - Metric for Evaluation of Translation with Explicit ORdering evaluates translations by aligning unigrams between the hypothesis and reference, computing a score based on precision, recall, and a fragmentation penalty.

Token \textbf{F1-Score} calculates the score based on the overlap of tokens between the predicted and reference sentences. It provides a balanced measure of token-level precision and recall. 


\section{Results} 
Model performance is presented in \autoref{tab:results}, for BLEU, METEOR, and F1, higher scores are better, while for WER and CER, lower scores indicate better performance. The results demonstrate that Hausa text correction is an achievable task, with the M2M100 (418M) model obtaining the best scores across all metrics, significantly restoring the noisy text to the original quality. 

Surprisingly, M2M100 performs better than models such as AfriTeVA, \texttt{UBC-NLP/cheetah-base}, and other models exposed to African languages. The model is on par with a much larger and more powerful model based on Llama 3 8B that has been instruction-tuned on over 400 million tokens of multilingual instruction data for African languages -- \texttt{NCAIR1/N-ATLaS}. The latter achieves the best overall scores ($\pm 0.001$ during multiple runs), however, given the much larger compute requirements of this model, we consider the best-performing model to be the smaller and more efficient M2M100.

Our experiments indicate that both the pretraining strategy and the architecture are important. AfriTeVA is a generic T5-like architecture pretrained on 10 African languages of which three languages dominate the corpus: Swahili, Tigrinya, and Hausa, languages that are from different language families: Bantu, Cushitic, Chadic. Cheetah is also a T5-like model, while M2M100 is an encoder–decoder model pretrained on a task much more similar to ours: many-to-many translations in 100 languages, including 18 African languages.

Furthermore, there is also the possibility of data contamination. The M2M100 model is pretrained on non-public data, which, according to \citet{fan2020englishcentricmultilingualmachinetranslation}, consists primarily of Common Crawl data and synthetic translations. The training data of AfriTeVA is the AfriBERTa corpus \cite{ogueji-etal-2021-small} which is also based on Common Crawl with more focus on quality. To test the data contamination hypothesis, all our evaluations exclude MinHashLSH approximate overlaps \cite{eric_zhu_2024_11462182} between AfriBERTa and our test set. We could only find 56 such overlaps at a 0.5 Jaccard coefficient threshold and their impact amounts in less than 0.001 changes in the final metrics. Additionally, we evaluate the models exclusively on the private OOD dataset (see \autoref{tab:ood_res}), which reveals an identical performance hierarchy among the models, albeit with lower overall scores.


Unsurprisingly, models lacking explicit Hausa pretraining such as mT5 and mBART perform subpar to any of the models pretrained on Hausa and African languages, with scores similar to GPT-5.1 and close to the Copy baseline.

\subsection{Human Evaluation}
We conduct a human evaluation of the best model's predictions using a set of 500 randomly selected genre-diverse samples. A native Hausa speaker is asked to provide binary assessments of correctness for the automatically corrected texts. The evaluation reveals that 358 predictions (71.6\%) are correct according to the reference. Out of the total correct predictions, 57 samples (12.5\%) contain at least one instance where hooked Hausa characters (ɓ, ɗ, ƙ, ƴ) are swapped with their Latin-looking counterparts. Furthermore, 135 samples (27\%) are labeled as incorrect. This incorrect label comprises instances where the model's output did not precisely match the reference word-for-word but the core meaning remained largely intact. Only a small fraction of 7 sentences (1.4\%) out of the total 500 were judged to have predictions that were worse than the original input.

A manual comparison between the M2M and N-ATLaS models reveals a very strong tie, henceforth, the human evaluation was focused more on AfriTeVA. These models, due to their pretraining task, tend to reconstruct the errors with paraphrasing and different words from the reference. For example, the model predicts "... wasannin gudun mita 100" (\textit{100m running competitions}) instead of the true "... tseren gudun mita 100" (\textit{100m race}). Semantically, the sentences may be similar, but this behavior is a source of downstream errors.

Appendix \ref{sec:app_samples} contains several examples of predicted texts from the best-performing model. The model tends to preserve the overall coherence of the phrase. As such, typos or misspellings that do not affect the coherence (i.e., a change of tense or person), but alter the original meaning are not modified. Furthermore, since the base model is multilingual, we observe several cases where named entities are contaminated from other languages - these are not corrected or transcribed using the standards for Hausa (e.g.,\textit{Pridnestrovie}), but using alternative spellings that are valid in other languages (e.g., \textit{Pridnistria}).

\section{Impact on Downstream Tasks}
To assess the practical impact of text correction, we conduct downstream evaluations on several tasks: text genre detection, machine translation, and LLM prompt response. These evaluations are designed to identify the extent to which noisy data degrades performance across tasks relative to the original data, and to determine whether automatic correction can recover the original performance levels.

\subsection{Genre Detection} 
For this task, we use the existing 750 chunks of the MultiplEYE out-of-domain dataset, translated and edited into Hausa by a professional translator.

\begin{figure}[!htb]
\centering
\includegraphics[width=0.8\linewidth]{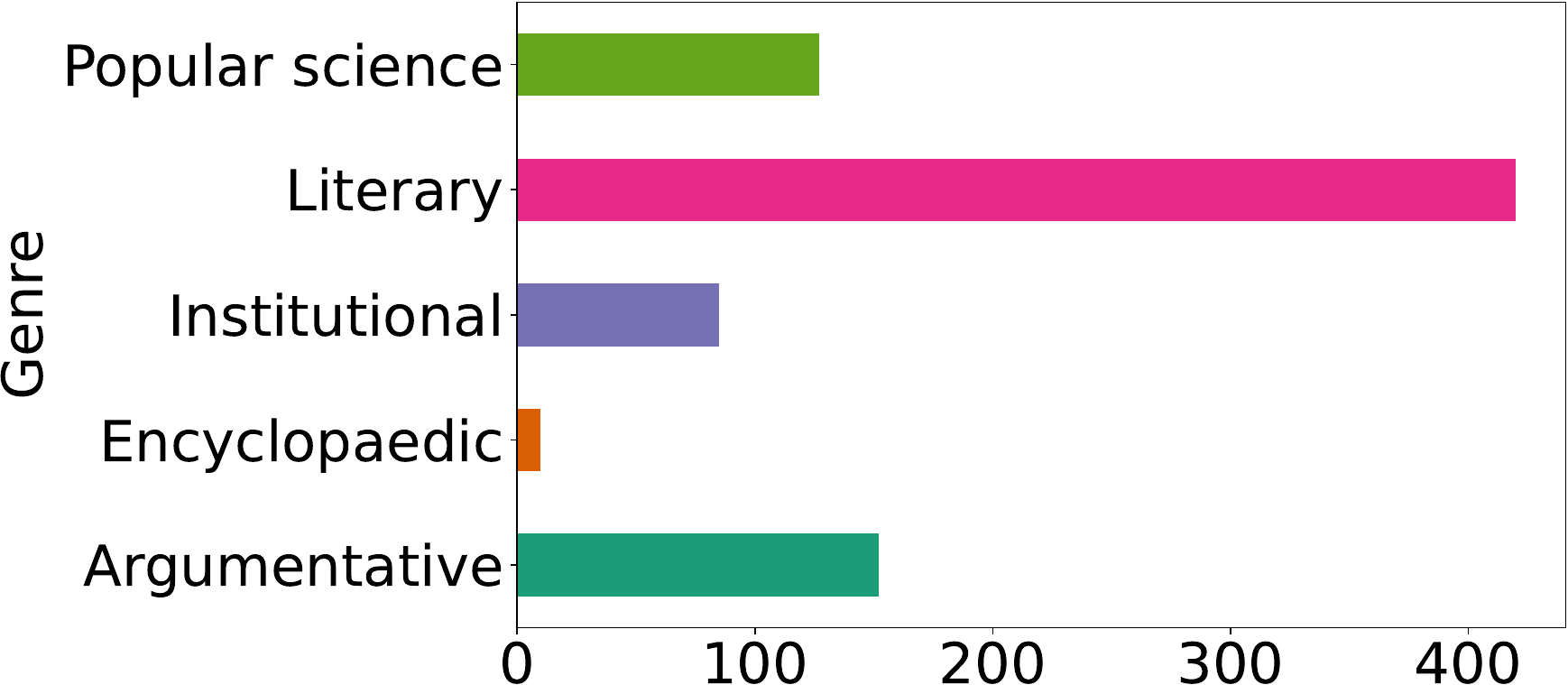}
\caption{Distribution of genres in the data.}
\label{fig:genres}
\end{figure}

The data contains two argumentative texts from the Programme for International Student Assessment (PISA), one text from Wikipedia, a part of the Universal Declaration of Human Rights, a text from the European Commission, seven literary texts, and two pop-science texts. The majority class consists of the Literary genre, as seen in \autoref{fig:genres}, and we do a stratified random split for model evaluation.

We train a genre-detection classifier at the sentence level based on the AfriBERTa-base\footnote{\url{https://huggingface.co/castorini/afriberta_base}} \cite{ogueji-etal-2021-small} model and evaluate the model using stratified 5-fold cross-validation. A comparative evaluation scores between classifiers trained on noise-induced data and automatically cleaned data and is displayed in \autoref{tab:genre_results}. 
\textbf{Original} represents human-edited, clean reference text (either from existing clean corpora or from professional translation).
\textbf{Noisy} represents texts obtained by injecting synthetic noise into the original text.
\textbf{Cleaned} are automatically corrected text produced by the normalization model when applied to the noisy input.

\begin{table}[htbp]
\centering
\begin{tabular}{lcc}
\toprule
Data & Accuracy & F1 \\
\midrule
\textbf{Original} & 80\% $\pm$ 2\% & 0.80 $\pm$ 0.02 \\
\textbf{Noisy} & 78\% $\pm$ 1\% & 0.77 $\pm$ 0.01 \\
\textbf{Cleaned} & 81\% $\pm$ 3\% & 0.80 $\pm$ 0.03 \\
\bottomrule
\end{tabular}
\caption{Genre detection performance comparison between models trained on noisy data versus cleaned data. Accuracy is slightly decreased on noised data while automatic correction recovers the original performance.
}
\label{tab:genre_results}
\end{table}

The noise in the data reduces the ability of the classifiers to correctly predict the genre, while the automatic cleaning restores the ability of models to learn and make predictions from the data, albeit with a slight increase in standard deviation.

We have experimented with the \texttt{mteb/NaijaSenti} sentiment dataset as well, without observing any statistically significant difference between the model on the noisy data vs. the model on the corrected dataset (78\% accuracy in both cases). The impact of noise on text classification is less stringent compared to other tasks, as we will see further.

\subsection{Hausa-to-English Machine Translation Task} 
Translation quality is evaluated using a subset of the FLORES+ benchmark \cite{nllb-24,wmt24-4african} consisting of 997-sentence human translated English-Hausa pairs. Noise was added using the same procedure as for the rest of the data using the parameters in \autoref{tab:noise_proba}. 

To translate, we use two proprietary models: Google Translate and GPT-5.1, and two open-source models: NLLB-200 \cite{nllb-24} and Tencent Hunyuan 7B \cite{hunyuan_mt}. The latter obtained the best evaluation scores across 30 out of 31 languages at the recent WMT2025 \cite{kocmi2025preliminaryrankingwmt25general}.
For LLM-based translation, the prompts used for the LLM-based translation systems are provided in \autoref{sec:LLM_prompt}.  
The evaluation is carried out using string based-metric BLEU and MetricX-24-Hybrid-XL \cite{juraska-etal-2024-metricx} and the results are rendered in \autoref{tab:translation_results}.

\begin{table}[htbp]
\centering
\resizebox{\columnwidth}{!}{%
\begin{tabular}{llccc}
\toprule
\textbf{System} & \textbf{Metric} & \textbf{Original} & \textbf{Noisy} & \textbf{Cleaned} \\
\midrule
\multirow{2}{*}{GPT-5.1}
& BLEU & 33.3 & 24.47 & 27.26 \\
& MetricX & 0.812 & 0.689 & 0.717 \\
\midrule
\multirow{2}{*}{Google Trans.}
& BLEU & 42.01 & 33.07 & 35.02 \\
& MetricX & 0.816 & 0.701 & 0.722 \\
\midrule
\multirow{2}{*}{NLLB-200}
& BLEU & 36.46 & 26.88 & 30.54 \\
& MetricX & 0.762 & 0.596 & 0.642 \\
\midrule
\multirow{2}{*}{Hunyuan}
& BLEU & 13.58 & 7.9 & 10.34 \\
& MetricX & 0.5 & 0.382 & 0.447 \\
\bottomrule
\end{tabular}
}
\caption{Translation quality comparison on original, noisy, and cleaned Hausa texts across four translation systems on the FLORES+ dataset. The automatically corrected texts consistently obtain better translations than the noisy texts. }
\label{tab:translation_results}
\end{table}

The best translation model, or rather the one that best matches the set FLORES+ is Google Translate, followed by the open-source NLLB-200, GPT-5.1 zero shot prompting, and Hunyuan MT. The latter has not been exposed to Hausa, therefore it is expected to perform worse.

A manual analysis of translations based on their similarity indicates that translations from cleaned texts are generally very similar to those from original texts, rather than closer to the reference. It is expected to observe lower evaluation scores of the cleaned sentences because the cleaned text is produced automatically and, while it removes orthographic noise, it may introduce unintended changes (see examples in Appendix \ref{sec:app_samples}). There are also artifacts in the FLORES+ dataset where the references are not a literal translation or cases where there is not enough context to translate properly the original (e.g., rows 226, 334, 495 in the development split\footnote{\url{https://huggingface.co/datasets/openlanguagedata/flores_plus}}).

For machine translation, unlike the text classification task, automatic recovery of errors is not perfect and phrases can lose their initial meaning and are therefore mistranslated into English.

\subsection{Instruction Following} 

\begin{table}[h]
\centering
\resizebox{1.02\columnwidth}{!}{%
\begin{tabular}{lccc}
\hline
\textbf{Task} & \textbf{Original} & \textbf{Noisy} & \textbf{Cleaned} \\
\hline
NER & 0.64 & 0.19 & 0.37 \\
\hline
AfriMMLU (all) & 79 & 74 & 76 \\
\quad Elem. Math & 82 & 79 & 78 \\
\quad Facts & 72 & 68 & 68 \\
\quad Geography & 82 & 73 & 77 \\
\quad Economics & 78 & 75 & 78 \\
\quad Law & 79 & 78 & 82 \\
\hline
AfriMGSM & 58 & 49 & 49 \\
AfriXNLI & 73 & 64 & 66 \\
\hline
\end{tabular}
}
\caption{Performance comparison across Original, Noisy, and Cleaned inputs. Named Entity Recognition is evaluated using the F1 score while all the others are evaluated using accuracy. NER is the most impacted task because it depends on identifying the exact word boundaries of entities.}
\label{tab:instruct_results}
\end{table}

This task is tested using a set of benchmarking data for  Hausa that evaluate several tasks such as Named Entity Recognition (MasakhaNER) \cite{hausaner} and the IrokoBench datasets recently published by \citet{adelani2024irokobench} consisting of Mathematical Reasoning QA (AfriMGSM), Natural Language Inference (AfriXNLI), and Multi-Choice QA (AfriMMLU). The latter is split by subject, such as elementary mathematics, global facts, high school geography, high school microeconomics, and international law.

We use the GPT-5.4 proprietary model and evaluate the zero-shot performance over the original, noised, and cleaned data, acknowledging that proprietary models may likely be contaminated by the data, especially since the benchmarks are translations of English variants. We evaluate NER using F1 score and all the other tasks using plain accuracy.

\autoref{tab:instruct_results} shows that proprietary large language models are sensitive to the noise introduced in the prompts and that a pre-processing step of cleaning and normalizing the data can lead to better instruction-following abilities. The named entity recognition (NER) task is the most affected (dropping from 0.64 original F1 score, 0.19 with noise, 0.37 cleaned) because the word boundaries are not preserved perfectly after cleaning up the noise. Despite this drop, the improvement over the noisy data is almost double.

For multiple-answer questions, i.e., Measuring Massive Multitask Language Understanding (MMLU), each subject has 100 questions and we can observe variation across subjects; for instance, geography and facts are more impacted than law or economics. Some of the results are incidental, such as law, where the cleaned version obtains higher scores than all the others, a result that raises some limitations of this type of evaluative comparison. Similarly, the drop in performance for Natural Language Understanding (NLI: 73 $\rightarrow$ 64 $\rightarrow$ 66) shows that text normalization is not able to restore the quality in order for commercial models to predict the correct textual entailment and that there is a surprisingly close gap between noised and corrected texts.

The African Multilingual Grade School Math Benchmark (MGSM) exhibits a substantial performance drop under noisy conditions, which is not recovered after cleaning. 
Our the findings indicate that noise robustness varies considerably by task, with structured prediction tasks being most vulnerable, and that automatic cleaning methods provide uneven gains depending on the nature of the task.

\section{Conclusions}
This paper addresses the prevalent issue of orthographic writing anomalies in digital Hausa text, which impedes the development of robust NLP tools. We compare several solutions centered on fine-tuning transformer models using a novel, synthetically-generated parallel corpus of over 400,000 noisy-to-clean Hausa sentence pairs. The dataset is designed to reflect realistic error patterns. 

Our experiments demonstrate that a sequence-to-sequence M2M100 (418M) model achieves a performance comparable to 8B models supervised-fine-tuned on Hausa and other African languages. A human and automatic evaluation of the methods also reveal the limitations of normalization and that the models may not be able to recover entirely the original data, which motivates future research directions for Hausa.

These findings show that smaller fine-tuned models, trained on synthetically noised data, can substantially enhance Hausa text quality and downstream NLP tasks including machine translation, text classification, and LLM instruction following. This has significant implications for Hausa NLP, as cleaner text resources are crucial to develop more accurate and reliable downstream applications such as machine translation and sentiment analysis, thereby improving digital resources for the Hausa-speaking community. This work provides a step towards enhancing Hausa language resources and offers a replicable methodology for tackling similar challenges in other low-resource languages.

To ensure full reproducibility, the synthetically generated noisy-clean parallel dataset, along with the code for our models, has been made public.\footnote{\url{https://github.com/ahmadmwali/HausaSeq2Seq}}

\section*{Limitations}
The current limitations of our work consist of
\begin{enumerate}
    \item the usage of social media data as a source of noise and user-generated anomalies; this noise is then projected onto good-quality edited texts
    \item the edited texts are from different genres and the synthetic noise might not be naturally occurring in such documents
    \item the evaluation of downstream tasks uses only automatic metrics which might introduce particular biases towards certain languages, further analysis is required to understand which parts of the text are irreparable and whether this is an actual issue in online texts
\end{enumerate}

\section*{Acknowledgments}
We would like to express our gratitude to Habib Sani Yahaya for his contribution to translating and revising the MultiplEYE texts.

This research is supported by InstRead: Research Instruments for the Text Complexity, Simplification and Readability Assessment  CNCS - UEFISCDI project number PN-IV-P2-2.1-TE-2023-2007 and by the project ``Romanian Hub for Artificial Intelligence - HRIA'', Smart Growth, Digitization and Financial Instruments Program, 2021-2027, MySMIS no. 351416.

\bibliography{custom}

\appendix

\section{Training Parameters}
\label{sec:app_training}

\paragraph{Hyperparameters for the T5, BART, and M2M models.}
Default training hyperparameters include a learning rate of 2e-5, a training batch size of 4 per device, an evaluation batch size of 8 per device, 3 training epochs, gradient accumulation steps of 4, and weight decay of 1e-3. The maximum sequence length for tokenization was set to 256 tokens, and generation during evaluation was configured with a maximum of 256 new tokens and 4 beams for beam search. Specific LoRA configurations involved a rank (r) of 16, alpha of 32, and dropout of 0.05.
Training was performed on NVIDIA L4 and H100 GPUs with the \textit{accelerate} library.

\paragraph{Hyperparameters for LLM fine-tuning}
The experiments use the Axolotl framework initialized from the \texttt{NCAIR1/N-ATLaS}. Training uses an instruction-formatted Hausa corpus derived from the \emph{noisy}-to-\emph{clean} correction task, reserving $5\%$ of the training file for validation. Model updates are performed with a micro-batch size of 1 and gradient accumulation steps of 32. The optimizer is an 8-bit AdamW from BitsAndBytes, configured with $\beta_1=0.9$, $\beta_2=0.999$, $\epsilon=1\times10^{-8}$, and a weight decay of $0.001$. The learning rate was set to $2\times10^{-5}$ using a cosine learning-rate scheduler with 100 warmup steps. Models were trained for two epochs, with evaluation conducted using batches of 8 samples and typical causal LM metrics (perplexity).

\onecolumn



\section{Data Samples}
\label{sec:app_samples}

Below we provide several examples of sentences that are automatically standardized using our best model. We have selected both perfect reconstructions and sentences with different types of errors. With boldface we highlight the main errors that make the prediction different from the true reference.

\begin{enumerate}[noitemsep, topsep=10pt, partopsep=0pt, parsep=0pt, itemsep=10pt]
    \item \textbf{Input (Noisy):} AnhaifiAbubakar Malami ,musulmin fulani , a ranar 17 gaAfrilunshekara ta 1967 a Birnin Kebbi, babban birnin JiharKebbi, Arewacin Najeria y .
    
    \textbf{Prediction:} An haifi Abubakar Malami , musulmin fulani , a ranar 17 ga Afrilun shekara ta 1967 a Birnin Kebbi , babban birnin Jihar Kebbi , Arewacin Najeriya .
    
    \textbf{Reference:} An haifi Abubakar Malami , musulmin fulani , a ranar 17 ga Afrilun shekara ta 1967 a Birnin Kebbi , babban birnin Jihar Kebbi , Arewacin Najeriya .

    \textbf{English translaiton:} Abubakar Malami, a Fulani Muslim, was born on 17th April, 1967, in Birnin Kebbi, the Kebbi State capital, Northern Nigeria.

    \textbf{Obs:} The prediction is identical to the reference, the original meaning is preserved. 

    \item \textbf{Input (Noisy):} Transnistria ( Prid nsetrovie ) , kuma anafurta Transdniestria , shi ne wani ɓangare na Moldova gabashin koginDniesterr kuma ( tun 1990 )a ka-ayyanadakuma fiyeko ƙasa da aiki zaman kanta a jihar ba ta re dda wani kasada kasa fitarwa da ga duk wanisarki jihar .

    \textbf{Prediction:} Transnistria ( \textbf{Pridnistria} ) , kuma ana furta \textbf{Transnistria} , shi ne wani ɓangare na Moldova gabashin kogin Dniester kuma ( tun 1990 ) \textbf{aka-yayya} da kuma fiye ko ƙasa da aiki zaman kanta a jihar ba tare da wani kasa da kasa fitarwa daga duk wani sarki jihar .

    \textbf{Reference:} Transnistria ( Pridnestrovie ) , kuma ana furta Transdniestria , shi ne wani ɓangare na Moldova gabashin kogin Dniester kuma ( tun 1990 ) a ka-ayyana da kuma fiye ko ƙasa da aiki zaman kanta a jihar ba tare da wani kasa da kasa fitarwa daga duk wani sarki jihar .

    \textbf{English translation:} Transnistria (Pridnestrovie), also pronounced Transdniestria, is a part of Moldova east of the Dniester River and (since 1990) has been declared as, more or less, an independent state without any international recognition from any sovereign state.

    \textbf{Obs:} The models makes mistakes at the named entity level, it uses a (valid) multilingual replacement (\textit{Pridnistria} from Macedonian) instead of the Hausa-specific variant (\textit{Pridnestrovie}). 

    \item \textbf{Input (Noisy):} Kwararrunna'ura nadan adam (HMI) shine tsarin sarrafa allo nadijital ( ta mafahii da fasahar Ha ptic ) wanda kuma a karaba tare ad SF90Stradale.

    \textbf{Prediction:} Kwararrun na 'ura na dan adam ( HMI ) shine tsarin sarrafa allo na dijital ( \textbf{ta mafi fiye da fasahar Hapttic} ) wanda kuma aka \textbf{karaba} tare da SF90 Stradale .

    \textbf{Reference:} Kwararrun na'ura na dan adam (HMI) shine tsarin sarrafa allo na dijital (ta amfani da fasahar Haptic ) wanda kuma aka raba tare da SF90 Stradale.

    \textbf{English translation:} Human Machine Interface (HMI) is a digital display control system (using Haptic technology) that is also shared with the SF90 Stradale.

    \textbf{Obs:} The original meaning is preserved, the model introduces a change of tense and a superlative expression.

    \item \textbf{Input (Noisy):} Dukena Courland ya kaf a sansaninSt.Andrea indaya sayi bayi kum aya say da kayan Ingilishi ,amma a ciki n Fabrairu 1660 ya sayar da wurin ga Dutch .

    \textbf{Prediction:} \textbf{Duk ne} na Courland ya kafa sansanin St. Andrea inda ya sayi bayi kuma ya sayar da kayan Ingilishi , amma a cikin Fabrairu 1660 ya sayar da wurin ga Dutch .
    
    \textbf{Reference:} Duke na Courland ya kafa sansanin St. Andrea inda ya sayi bayi kuma ya sayar da kayan Ingilishi , amma a cikin Fabrairu 1660 ya sayar da wurin ga Dutch .
    
    \textbf{English translation:} The Duke of Courland established the St. Andrea fort where he bought slaves and sold English goods, but in February 1660 he sold the place to the Dutch.

    \textbf{Obs:} The model changes the subject of the sentence (``Duke of Courland'' with ``All of Courland'', mistaking ``Duke'' with the Hausa word ``duk'' that translates to ``all'' ) making the sentence ungrammatical and difficult to read.

    \item \textbf{Input (Noisy):} Hydrogen chloride, acetic acid , da mai yawansauran Brønsted-Lowry aci d ba zasu iya samar da haɗin kai tare da nau 'in electron ba , duk da haka , don haka ba Lewis acid ba ne .
    
    \textbf{Prediction:} Hy ( Sin , acetic acid , da mai yawan wasa na (n ƙungiya-L purposery aci da ba za su iya samar da haɗin kai tare da nau 'in , duk da haka , don ba haka ba acid ba ne .

    \textbf{Reference:} Hydrogen chloride , acetic acid , da mafi yawan sauran Brønsted-Lowry acid ba za su iya samar da haɗin kai tare da nau 'in electron ba , duk da haka , don haka ba Lewis acid ba ne . 

    \textbf{English translation:} Hydrogen chloride, acetic acid, and most other Brønsted-Lowry acids cannot form coordination bonds with electron pairs, therefore, they are not Lewis acids.
    
    \textbf{Obs:} The meaning has changed from the original. Although the input is quite understandable, the model's prediction becomes completely incoherent (the model's prediction roughly translates to ``Hy (China, acetic acid, and the very playful of the (n L-team purposery to eat with the possibility of providing cooperation together with the type, even though, therefore it is not an acid.'') 




    
\end{enumerate}

\section{GPT Zero Shot Correction Prompt}
\label{sec:LLM_correction_prompt}

You will be given a sentence in Hausa and your task is to automatically correct it. Please respect the following indications:
\begin{enumerate}[noitemsep, topsep=0pt, partopsep=0pt, parsep=0pt, itemsep=0pt]
   \item    Read the sentence carefully, analyze it to understand it's context.
\item    Correct any grammatical, spelling, or punctuation errors in the sentence.
\item    Ensure that the corrected sentence maintains the original meaning and context.
\item    Provide the corrected sentence as your output without any additional explanations or comments.
\item    Do not change the structure of the sentence more than necessary to correct the errors.  
\end{enumerate}

\section{LLM SFT Instruction}
\label{sec:llm_sft}
\#\#\# Instruction:

Correct the noisy Hausa text below into a clean, grammatically correct version. Do not change the structure of the sentence.

\#\#\# Input:

\{noisy\}

\#\#\# Response:

\section{LLM Translation Prompt}
\label{sec:LLM_prompt}

You are a professional translator specializing in Hausa to English translation. You will receive text in Hausa that may contain informal language, colloquialisms, and errors.

Your task: \begin{enumerate}[noitemsep, topsep=0pt, partopsep=0pt, parsep=0pt, itemsep=0pt]
\item Translate the provided Hausa text to English accurately
\item Preserve the meaning and tone of the original text
\item  For "noisy input": Translate as-is, maintaining any informal language or errors in meaning but translate them to English
\item For "clean input": Translate the clean, corrected version naturally

\end{enumerate}

Guidelines:
\begin{itemize}[noitemsep, topsep=0pt, partopsep=0pt, parsep=0pt, itemsep=0pt]
\item Maintain natural English flow while staying faithful to the Hausa original
\item Keep cultural context and idioms where possible
\item If text contains code-switching, translate only the Hausa portions
\item Do not add explanations or notes, only provide the translation
\item Preserve any numbers, names, or specific terms as they appear

\end{itemize}

\end{document}